\documentclass[10pt,twocolumn,letterpaper]{article}

\usepackage{iccv}
\usepackage{times}
\usepackage{epsfig}
\usepackage{graphicx}
\usepackage{amsmath}
\usepackage{amssymb}

\usepackage{booktabs}

\usepackage{multirow}
\usepackage{bm}
\usepackage{color}
\usepackage[table,xcdraw,dvipsnames]{xcolor}
\definecolor{darkspringgreen}{rgb}{0.0, 0.45, 0.05}
\definecolor{carmine}{rgb}{0.68, 0.05, 0.0}
\definecolor{iODBlue}{RGB}{220, 232, 250}
\usepackage{pifont}

\newcommand{\cmark}{\ding{51}}%
\newcommand{\xmark}{\ding{55}}%

\usepackage[breaklinks=true,bookmarks=false]{hyperref}

\usepackage[accsupp]{axessibility}

\iccvfinalcopy 

\def\iccvPaperID{6578} 

\ificcvfinal\fi

\begin{document}

\title{GrowCLIP: Data-aware Automatic Model Growing for \\ Large-scale Contrastive Language-Image Pre-training}



\author{Xinchi Deng\textsuperscript{1} \quad Han Shi\textsuperscript{2} \quad Runhui Huang\textsuperscript{1} \quad Changlin Li\textsuperscript{3} \quad Hang Xu\textsuperscript{2} \\ \quad Jianhua Han\textsuperscript{2} \quad James Kwok\textsuperscript{4} \quad Shen Zhao\textsuperscript{1}\thanks{Corresponding author.} \quad Wei Zhang\textsuperscript{2} \quad Xiaodan Liang\textsuperscript{1} \\
{\normalsize
\textsuperscript{1}Sun Yat-sen University \quad \textsuperscript{2}Huawei Noah's Ark Lab} \\ {\normalsize \quad \textsuperscript{3}University of Technology Sydney \quad \textsuperscript{4}The Hong Kong University of Science and Technology}\\
{\tt\small dengxch5@mail2.sysu.edu.cn}\\
}

\maketitle
\ificcvfinal\thispagestyle{empty}\fi

\begin{abstract}
Cross-modal pre-training has shown impressive performance on a wide range of downstream tasks, benefiting from massive image-text pairs collected from the Internet. 
In practice, online data are growing constantly, highlighting the importance of the ability of pre-trained model to learn from data that is continuously growing.
Existing works on cross-modal pre-training mainly focus on training a network with fixed architecture.
However, it is impractical to limit the model capacity when considering the continuously growing nature of pre-training data in real-world applications.
On the other hand, it is important to utilize the knowledge in the current model to obtain efficient training and better performance.
To address the above issues, in this paper, we propose GrowCLIP, a data-driven automatic model growing algorithm for contrastive language-image pre-training with continuous image-text pairs as input. 
Specially, we adopt a dynamic growth space and seek out the optimal architecture at each growth step to adapt to online learning scenarios.
And the shared encoder is proposed in our growth space to enhance the degree of cross-modal fusion.
Besides, we explore the effect of growth in different dimensions, which could provide future references for the design of cross-modal model architecture.
Finally, we employ parameter inheriting with momentum (PIM) to maintain the previous knowledge and address the issue of the local minimum dilemma. Compared with the existing methods, GrowCLIP improves $2.3\%$ average top-$1$ accuracy on zero-shot image classification of  $9$ downstream tasks. As for zero-shot image retrieval, GrowCLIP can improve $1.2\%$ for top-$1$ image-to-text recall on Flickr30K dataset.
\end{abstract}

\section{Introduction} \label{sec:intro}

Recently, large-scale pre-trained models illustrate the potential performance among different fields including computer vision (CV)~\cite{he2021masked,he2020momentum,chen2020simple} and natural language processing (NLP) \cite{devlin2019bert,liu2019roberta,lan2019albert}. Cross-modal models like CLIP \cite{radford2021learning}, ALIGN \cite{jia2021scaling}, FILIP \cite{yao2022filip}, BLIP \cite{li2022blip} also demonstrate remarkable success across various vision-language downstream tasks. Note that these models are usually built in dual-stream architectures, which consist of the image and text encoders to extract the feature of image and text inputs, respectively. The alignment between these two features is then performed under the contrastive objective to learn the alignment between different modals.

\begin{figure}
\begin{center}
\includegraphics[width=1\linewidth]{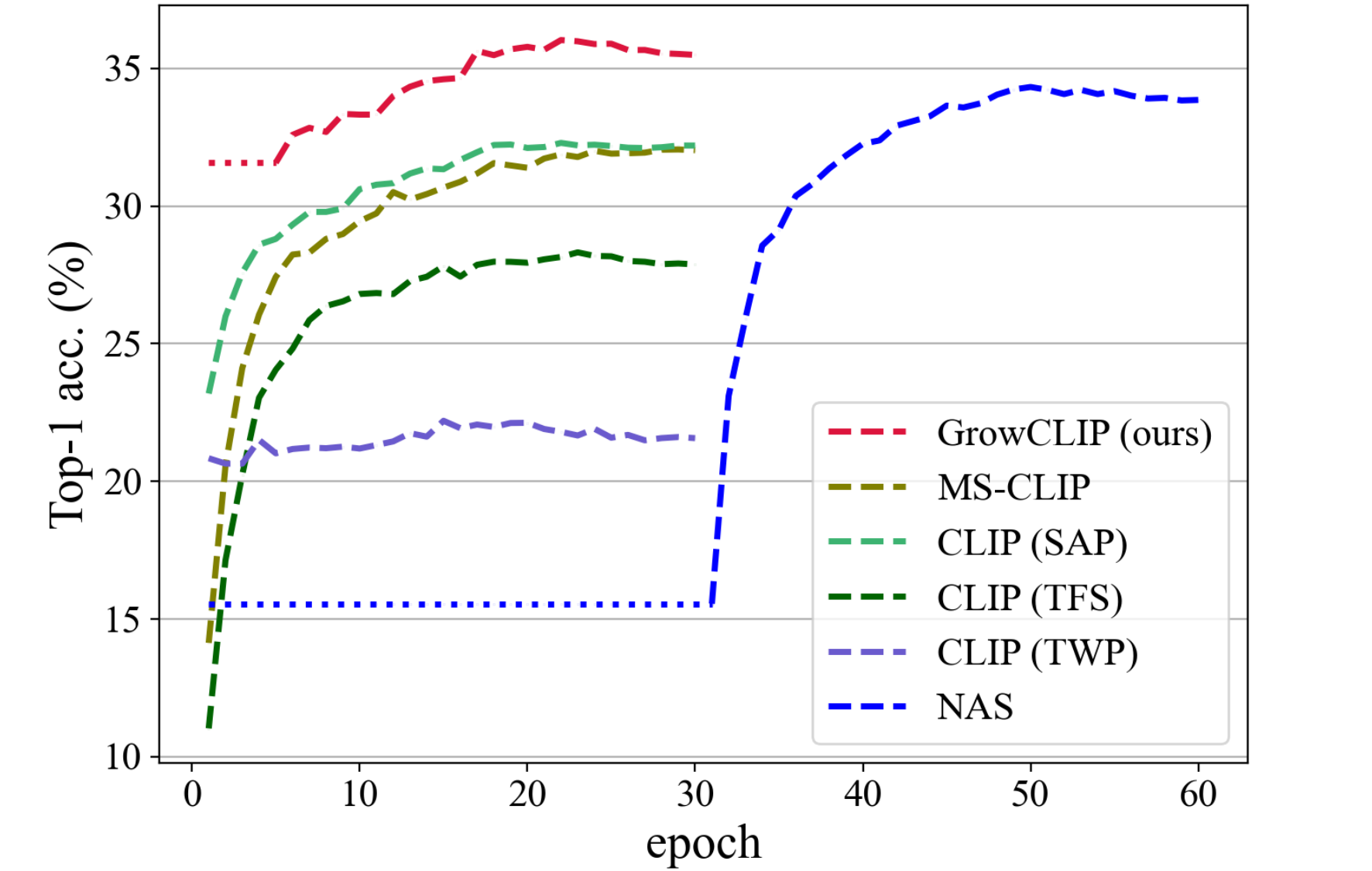}
\end{center}
\vspace{-1em}
   \caption{Top-1 accuracy (\%) of zero-shot image classification on ImageNet of GrowCLIP and baselines during training at step $4$, where the horizontal dotted lines mean the process of supernet training. Our GrowCLIP has the best performance and is more efficient compared with other baselines.} 
\label{fig:1}

\end{figure}

However, training these cross-modal models requires a large amount of image-text pairs collected from the Internet, 
e.g., $400$M image-text pairs for CLIP~\cite{radford2021learning} and $300$M for FILIP~\cite{yao2022filip}.
Most existing cross-modal methods \cite{radford2021learning,jia2021scaling,yao2022filip,li2021supervision,li2022blip,yu2022coca,wang2023boosting,alayracflamingo} directly train the final large-scale model using the completed image-text pair datasets collected at some point in time.
However, it is impractical to regard cross-modal pre-training as ``disposable'' training without considering the continuously growing nature of pre-training data or domain in real-world applications. 
For instance, mountains of new knowledge is generated on the Internet every day, which can be used to further improve the performance of existing pre-trained cross-modal models.
Therefore, in this paper, we consider the training of large-scale cross-modal pre-trained models as an online learning problem \cite{grabner2006line,saffari2009line,shalev2012online} with continuous image-text pairs as input.
Different from the standard continual learning setting  \cite{srinivasanclimb,yan2022generative,hadsell2020embracing,kirkpatrick2017overcoming,aljundi2018memory,buzzega2020dark}, the previous image-text pairs can also be achieved in our setting since we assume the size of training data is gradually increased over time.

One tough problem in the online learning scenario is that the model capacity should be related to the size of training data \cite{dosovitskiy2021image,zhai2022scaling,kaplan2020scaling,hu2021scaling}.
For example, it's observed that 
large ViT models perform worse than ResNets when pre-trained on small datasets, while the result is opposite when they are both pre-trained on larger datasets
\cite{dosovitskiy2021image}.
To verify the relationship between model capability and data size, we split Conceptual 12M (CC12M) dataset \cite{changpinyo2021cc12m} and test the performance of ViT-B/16 with different scales \cite{radford2021learning}.
As shown in Table~\ref{tab:1}, the relative performance ranking is dependent on the size of the training data. Given a smaller dataset (e.g., $10\%$ CC12M), the performance of 0.5-ViT-B/16 is comparable with ViT-B/16. In contrast, when data is sufficient (e.g., $100\%$ CC12M), the performance of ViT-B/16 is better than the smaller model. Thus, the unchanged model is not practical for this real-time learning setting and it's still an open issue on how to modify and train our model with the growing data. 
Besides, how to efficiently make use of knowledge of previous model when new data is coming remains an open problem.
One direct solution is to fine-tune the model with the updated training dataset.
However, training with previous pre-trained parameters of the same model will deteriorate the performance  \cite{ash2020warm} due to the parameter inheriting issue. 
As shown in Figure~\ref{fig:1}, the CLIP training with pre-training (TWP) have worse performance than the one training from scratch (TFS).

\begin{table}
\begin{center}
\begin{tabular}{lcccc}
\toprule
CC12M &
$10\%$ &
$20\%$&
$50\%$ &
$100\%$ \\
\midrule
CLIP-0.5-ViT-B/16 & 6.6 & 11.1 & 19.0 & 25.7  \\
CLIP-ViT-B/16 & 6.5  & 11.6 & 20.9 & 28.3  \\
\bottomrule
\end{tabular}
\end{center}
\vspace{-2mm}
\caption{Result of different model sizes with different data sizes: Top-1 accuracy (\%) of zero-shot image classification on ImageNet. The depth and the width of 0.5-ViT-B/16 are both one half of ViT-B/16. The relative performance ranking is dependent on the size of the training data.}
\label{tab:1}
\vspace{-6mm}
\end{table}

To address the above issues, we propose a data-aware automatic model growing method (denoted as GrowCLIP) for large-scale contrastive language-image pre-training, which performs a model growth process considering the gradually increased pre-training data. 
Specially, 
when training data grows dynamically, we adopt different cross-modal network architectures via the customized neural architecture search~(NAS) to make the network pre-training more efficient. Different from traditional NAS approaches \cite{zoph2018learning,real2018regularized,xie2018snas}, we introduce a cross-modal customized NAS by defining a dynamic search space named growth space, which is enlarged when more data comes, and proposing a shared encoder search space to enhance the degree of cross-modal fusion.
To utilize the architecture at the previous step more efficiently, the parameters of the new architecture are also inherited from the old one with momentum to maintain the previous knowledge and address the issue of local minimum dilemma.
Finally, growth architecture selection procedure is performed to select the optimal model architecture at each step, considering the performance and model size.

Experiments are conducted by averagely dividing the Conceptual 12M (CC12M) into $4$ growth steps under the online learning setting. As depicted in Table~\ref{fig:1}, compared with the existing methods, our GrowCLIP has the best performance and is more efficient.
Specially, experimental results show that our GrowCLIP can improve up to $2.3\%$ average top-1 accuracy on zero-shot image classification of 9 downstream tasks compared with the existing methods. As for zero-shot image-text retrieval, GrowCLIP has a $1.2\%$ improvement for top-$1$ image-to-text recall on  Flickr30K \cite{plummer2015flickr30k} dataset and $0.8\%$ on MSCOCO \cite{lin2014microsoft}. 

To summarize, the contributions of this paper are listed as follows: (i) To adapt to the growing data scenario, we propose a data-aware automatic model growing method, named GrowCLIP. (ii) We provide some insights for the design of cross-modal model architecture. (iii) The extensive experiment results illustrate the effectiveness of our approach on zero-shot classification and retrieval tasks.

\section{Related Work}
\textbf{Vision-Language Pre-training Models} \; Inspired by the success of pre-training in computer vision (CV) and natural language processing (NLP), a boosting number of research works in the domain of vision-language pre-training (VLP) has recently surged to pursue a unified multi-modality representation.
Vision-language pre-training trains the model on large-scale image-text pairs to improve performance
of downstream vision and language tasks including image classification \cite{deng2009large,fei2006one,krizhevsky2009learning}, retrieval \cite{plummer2015flickr30k,lin2014microsoft}, Image
Captioning (IC) \cite{liu2021cptr,cornia2019show,hu2020vivo,hu2021scaling}, Visual Question and Answer (VQA) \cite{antol2015vqa,lu202012,yu2021ernie} and so on.
The VLP model can be categorized into two categories: (i) The dual-stream model, e.g., CLIP \cite{radford2021learning}, ALIGN \cite{jia2021scaling}, FILIP \cite{yao2022filip}, processes visual and textual tokens separately with two parallel streams to acquire the representation and then fuse them through interactive module. (ii) The single-stream model, e.g., Oscar \cite{li2020oscar}, ViLT \cite{kim2021vilt}, VisualBERT \cite{li2019visualbert}, directly combines the extracted visual and textual tokens and feeds them into the transformer-based model. The public pre-training datasets of VLP model include  available datasets like YFCC100M \cite{thomee2016yfcc100m}  and CC12M \cite{changpinyo2021cc12m}. CLIP \cite{radford2021learning}, ALIGN \cite{jia2021scaling}, FILIP \cite{yao2022filip} and LEMON \cite{hu2021scaling} adopt large-scale image-text pairs crawled from the web to show more powerful models. 
Different from these works that treat this process as offline learning, we investigate the online learning case in vision-language pre-training.

\textbf{Online Learning} \;Online learning is a problem setting of machine learning in which data becomes available in a sequential order \cite{grabner2006line,saffari2009line,shalev2012online}. Different from continual learning \cite{hadsell2020embracing, kirkpatrick2017overcoming, aljundi2018memory, buzzega2020dark}, online learning can reach previous data all the time. As opposed to batch learning techniques which generate the best model by learning on the entire training dataset at once, online learning is used to update our predictor for future data at each growth step.
Follow the leader \cite{song1989study} is the simplest learning rule which assumes the hypothesis that the leader has the least loss overall past rounds. Follow the regularised leader \cite{pogodin2020first} is a modification variant and obtain better regret bounds by learning a regularisation function. 
Online convex optimization \cite{hazan2016introduction} is a general algorithm framework for decision making which targets convex optimization. To adapt online learning scenario, we propose a data-aware automatic model growing algorithm that grows the model with increasing training data.

\textbf{Neural Architecture Search} \;Neural Architecture Search (NAS) aims to automatically design the state-of-the-art neural networks. Early works \cite{zoph2018learning,real2017large,tan2019mnasnet} are based on reinforcement learning and evolutionary algorithms. In the later work, various techniques are used to reduce the computational complexity. Gradient-based methods \cite{liu2018darts,liu2018progressive} regard the network architecture as a set of parameters and adopt back-propagation algorithm to optimize them. One-shot NAS methods \cite{peng2020cream,chen2021autoformer,guo2020single} train a supernet to avoid training each subnet from scratch. Unfortunately, one-shot NAS methods have to spend lots of resources to train the supernet. In contrast, the method we propose inherits the parameters of the old model with momentum, which only takes few epochs to train the supernet. Furthermore, GrowCLIP can modify the neural architecture to adapt to the size of the training data.

\section{Methodology}

\begin{figure*}
\begin{center}
\includegraphics[width=1\linewidth]{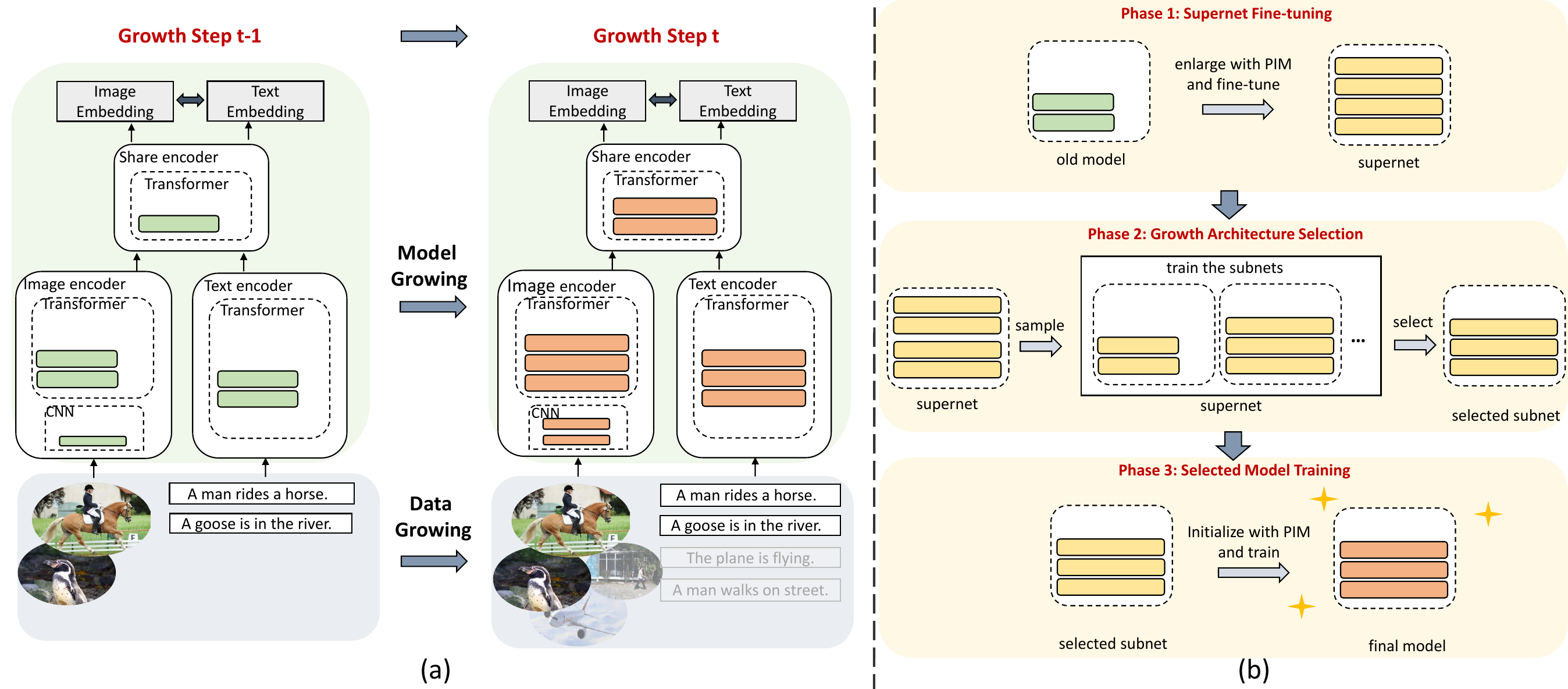}
\end{center}
   \caption{The overview of GrowCLIP. As shown in Figure (a), when the data grows from growth step $t-1$ to growth step $t$, our model grows adaptively as well. Figure (b) illustrates the detail of model growing and we simplify the representation of model architecture.}
\label{fig:main}
\end{figure*}
In this section, we describe the online learning problem setting and its challenges for cross-modal pre-training (Section~\ref{sec:setting}). To adapt the growth data, we propose a dynamic growth space, including a proposed shared encoder (Section~\ref{sec:space}). To rise to the challenge of the local minimum dilemma, we propose parameter inheriting with momentum (Section~\ref{sec:pim}). The whole pipeline of our method is illustrated in Section~\ref{sec:pipeline}.

\subsection{Problem Setting and Challenges} \label{sec:setting}
\textbf{Problem Setting} \; The target of online learning is obtaining the optimal model at each growth step of the data. Let $\mathcal{D}^{t-1}=\{\bm{x}^I_i,\bm{x}^T_i\}_{i=1}^k$ be the training data at growth step $t-1$, where $\bm{x}^I$ and $\bm{x}^T$ are image and text samples and $k$ is the training data size. At the next growth step $t$, the enlarged training dataset $\mathcal{D}^{t}$ satisfies $\mathcal{D}^{t-1} \subseteq \mathcal{D}^{t}$ due to the incoming data. Given an optimal model from growth step $t-1$, our aim is to achieve the optimal model (with optimal architecture $\bm\psi^*$ and parameters $\bm\omega^*$) based on the enlarged dataset $\mathcal{D}^{t}$:

\begin{equation}
    \{\bm\psi^*, \bm\omega^*\} = \mathop{\arg\min}_{\bm\psi, \bm\omega}\mathcal{L}(\bm\psi, \bm\omega; \mathcal{D}^{t}).
\end{equation}

\textbf{Challenge of Data Growth} \; As discussed in Table~\ref{tab:1}, the size of the training dataset is related to the optimal architecture selection. Given a smaller dataset, the small-scale architecture has comparable performance to large-scale architecture but fewer parameters are deployed. When data grows, the small-scale architecture may limit the final performance due to the limited parameters. However, few works notice the effect of data size in online learning settings.

\textbf{Challenge of Parameter Inheriting} \; As investigated, the performance of model trained from scratch is much better than trained with pre-training. This is probably caused by the influence of the inheriting parameters trained with the previous smaller dataset, which makes it difficult for the model to escape from the local minima in the grown dataset. However, training from scratch is costly due to multiple growth steps in online learning. The balance between training from scratch and training with pre-training is worthy of consideration.

To solve the above problems, we propose an automatic model growing method - GrowCLIP (shown in Figure~\ref{fig:main}), which can modify the neural architecture at different growth steps in online learning case.
To adapt to the growing data scenario, we propose a growth space for each growth step,  where the scale of architecture candidates among this space is larger when data grows. 
To alleviate the parameter inheriting  issue with efficiency, we perform parameter inheriting with momentum such that the trade off between training from scratch and training with pre-training is balanced.

\subsection{Growth Space} \label{sec:space}
\textbf{Model Architecture}\; Following the basic neural architecture in CLIP \cite{radford2021learning}, we adopt a dual-stream model with image encoder $f(\cdot)$ and text encoder $g(\cdot)$. To adapt the cross-modality scenarios, we propose a shared encoder $h(\cdot)$, in which the image and text share the transformers except layernorm layers. The shared encoder not only can reduce model parameters, but also  enhance the degree of cross-modal fusion.
Given an image $\bm{x}^I_i$ and a text $\bm{x}^T_j$,
the model outputs the image representations $I_i = h(f(\bm{x}^I_i))$  and text representations $T_j = h(g(\bm{x}^T_j))$.
The similarity score computation of between image $\bm{x}^I_i$ and text $\bm{x}^T_j$ is:
\begin{equation}
  s_{i,j} =  h(f(\bm{x}^I_i))^\top h(g(\bm{x}^T_j)),
\end{equation}
where $s_{i,j}$ denotes the similarity of the $i$-th image to the $j$-th text. 

In each training batch, we sample $n$ image-text pairs $\{\bm{x}^I_i, \bm{x}^T_i\}_{i=1}^n$ from the dataset.
For image $\bm{x}^I_i$ in the image-text pair $<\bm{x}^I_i, \bm{x}^T_i>$, $\bm{x}^T_i$ is its positive, while the other texts will be used as in-batch negatives. 
The image-to-text  contrastive loss $\mathcal{L}^I_i$ for $\bm{x}^I_i$  can be formulated as:
\begin{equation}
\mathcal{L}^I_i \big(\bm{x}^I_i, \{\bm{x}^T_j\}_{j=1}^n\big) = -\frac{1}{n} \log \frac{\exp(s_{i,i})}{\sum_{j}\exp(s_{i,j})},
\end{equation}
Similarly, the text-to-image contrastive loss for $\bm{x}^T_i$ is:
\begin{equation}
\mathcal{L}^T_i \big(\bm{x}^T_i, \{\bm{x}^I_j\}_{j=1}^n\big) = -\frac{1}{n} \log \frac{\exp(s_{i,i})}{\sum_{j} \exp(s_{i,j})},
\end{equation}
Totally, the contrastive loss of this mini-batch can be represented by $\mathcal{L} = \frac{1}{2}\sum_{i=1}^a \big(\mathcal{L}^I_i + \mathcal{L}^T_i\big)$.

\textbf{Growth Space of image encoder} \; For the image encoder, we adopt a  mixed architecture of convolutional layer and transformer block \cite{vaswani2017attention}. There are three variable factors in our growth space design: (i) the number of convolutional layers $l^I$; (ii) the number of transformer blocks $b^I$; (iii) the number of transformer heads $h^I$.

\textbf{Growth Space of text encoder} \; The text encoder part is a transformer-based architecture, whose growth space includes two variable factors: (i) the number of transformer blocks $b^T$ and (ii) the number of transformer heads $h^T$. 

\textbf{Growth Space of shared encoder} \; The shared encoder is a transformer-based architecture, whose growth space includes the number of transformer blocks $b^S$.

\begin{table}[h]
\begin{center}
\setlength{\tabcolsep}{3.3pt}
 \resizebox{0.48\textwidth}{!}{
\begin{tabular}{lccc}
\toprule
&Image encoder &
Text encoder&
Shared encoder\\
\midrule
Transformer blocks & \{$b_{t-1}^I$, $b_{t-1}^I+4$\} & \{$b_{t-1}^T$, $b_{t-1}^T+4$\} & \{$b_{t-1}^S$, $b_{t-1}^S+4$\}   \\
Transformer heads & \{$h_{t-1}^I$, $h_{t-1}^I+4$\} & \{$h_{t-1}^T$, $h_{t-1}^T+4$\} & \textcolor{gray}{-}    \\
Convolutional layers & \{$l_{t-1}^I$, $l_{t-1}^I+2$\} &   \textcolor{gray}{-}
&   \textcolor{gray}{-}\\ 
\bottomrule
\end{tabular}}
\end{center}
\caption{The growth space $\Phi$ of GrowCLIP at the growth step t. $b_{t-1}^I$, $b_{t-1}^T$, $b_{t-1}^S$, $h_{t-1}^I$, $h_{t-1}^T$, $l_{t-1}^I$ are the hyperparameters of model architecture at growth step $t-1$.}
\label{tab:growth space}
\end{table}

The detailed descriptions of growth space for three encoders are shown in Table~\ref{tab:growth space}. At each new growth step, the variable factors are enlarged to adapt the growth data. Specifically, the numbers of transformer blocks and transformer heads are both increased by $4$, and the number of convolutional layers is increased by $2$.

\subsection{Parameter inheriting with momentum} \label{sec:pim}

\begin{figure}[t]
\begin{center}
\includegraphics[width=1\linewidth]{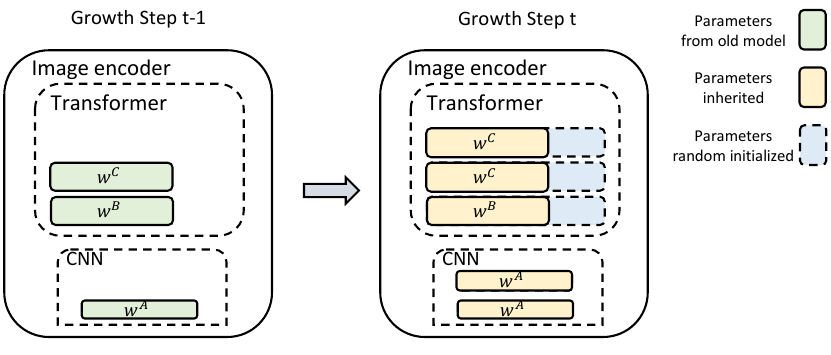}
\vspace{-9mm}
\end{center}
   \caption{Take image encoder for example of our proposed parameter inheriting with momentum. As for the initialization of the enlarged model, we preserve the existing parameters of the old model for exploitation and combine it with random initialization for exploration.}
\label{fig:para_inherit}
\vspace{-5mm}
\end{figure}

Some previous papers utilize the training results of the old model and transfer the knowledge to the new model \cite{grill2020bootstrap,guo2020bootstrap,he2020momentum,ash2020warm,li2022automated}.  
Inspired by the above methods, we propose parameter inheriting with momentum (PIM) to avoid local minimum dilemma. 
The illustration of our parameter inheriting with momentum is shown in Figure~\ref{fig:para_inherit}.
To preserve the knowledge from the old model, we extend the size of the old model to the enlarged model and maintain the parameters in existing parts.
As for the extended layers, we copy the parameters from the last layers directly. $\bm\omega_\textit{old}$ denotes the parameters of extension of the old model.
Besides the above parameter inheriting term, we also add a re-initialization term  with randomly initialized parameters  $\bm\omega_\textit{rand}$ to avoid the parameter inheriting issue. The final initialized supernet is a combination of the above two terms as follows:
\begin{equation}
\bm\omega_\textit{inherit} = \beta\cdot\bm\omega_\textit{old} + \gamma\cdot\bm\omega_\textit{rand},
\end{equation}
where $\beta$ and $\gamma$ are two hyperparameters related to exploitation and exploration. Additionally, the parameters of new transformer heads are randomly initialized, which makes that different heads in transformer are orthogonal and the new heads learn new features from different perspectives.

\subsection{Growth Pipeline} \label{sec:pipeline}
The growth pipeline at growth step $t$ (except the first one) includes three sequential phases as shown in Figure~\ref{fig:main} (b) . 

\textbf{Phase 1: Supernet Fine-tuning} \; 
As for a new growth step, a supernet $\Psi$ covering the whole growth space is constructed by enlarging the current model.
First, we use the PIM method described in Section~\ref{sec:pim} to initialize the supernet. However, there are some newly added layers in the supernet compared with the old model from the last growth step, which brings a performance gap between the subnets with new layers and others. To make the supernet more robust, we slightly fine-tune the parameters of the full supernet for a few epochs to mitigate the performance gap among subnetworks after PIM initialization.

\textbf{Phase 2: Growth Architecture Selection} \;
Growth Architecture Selection (GAS) is the main phase in our algorithm
where we select the optimal architecture in the growth space. The whole GAS process can be divided into two steps: training and searching.

\textit{First}, to find the optimal architecture $\bm\psi^*$, we need to fully optimize all the subnet in the supernet, which covers all the candidate growth architecture:
\begin{equation}
\label{eqn:supernet_param}
\bm\omega_\textit{super}^* = \mathop{\arg\min}_{\bm\omega_\textit{super}}  \mathbb{E}_{\bm\psi\in\Psi}~\{\mathcal{L}(\bm\psi,\bm\omega_\textit{super}; \mathcal{D}^{t})\}.
\end{equation}
Inspired by one-shot neural architecture search \cite{cai2019once}, we optimize $\bm\omega_\textit{super}$ in Equation~\ref{eqn:supernet_param} by uniformly sampling subnets $\bm\psi$ from the growth space $\Phi$ and update the parameters of the sampled subnets in each iteration.
More specifically, at each step, we first randomly sample a subnet $\bm\psi$ from the growth space $\Psi$, then optimize the parameter of this subnet by minimizing the contrastive loss between image and text. This can be considered as solving Equation~\ref{eqn:supernet_param} via Monte Carlo sampling. By iterative training, these subnets can be fully optimized in a few epochs.

\textit{Second}, after training the supernet, we search for the optimal subnet in the growth space $\Phi$.
Here, we consider both the performance and the number of parameters of the architecture,
and select the optimal architecture $\bm\psi^*$ using the composite metric: 
\begin{equation} \label{eq:metric}
\bm\psi^* = \mathop{\arg\max}_{\bm\psi \in \Psi}\left\{ \mathcal{A}(\bm\psi) + \alpha\frac{\mathcal{N}^{t-1}}{\mathcal{N}^t} \frac{\mathcal{P}(\Psi)}{\mathcal{P}(\bm\psi)}\right\},
\end{equation}
where $\mathcal{A}(\cdot)$ is the top-$1$ accuracy of test dataset, $\mathcal{N}^t$ denotes the number of data on growth step $t$, $\mathcal{P}(\cdot)$ represents the total amount of model parameters and $\alpha$ is the trade-off hyperparameter. 
When we get a large number of new coming data, $ \frac{\mathcal{N}^{t-1}}{\mathcal{N}^t}$ will get small, which means that we focus more on the performance and encourage the growth of models.
As our growth space is relatively small, traversal evaluation of all the candidate architectures is affordable. Therefore, we simply test all candidate subnets in the growth space $\Phi$ using the parameters inherited from the supernet $\Psi$ and select the optimal one for the next stage via the composite metric of Equation~\ref{eq:metric}.

\textbf{Phase 3: Selected Model Training} \;
To alleviate the influence of supernet training on the grown model, we also use PIM to initialize the parameter $\bm\omega$ of selected model $\bm\psi^*$:
\begin{equation}
\bm\omega = \beta\cdot\bm\omega^*_\textit{super} + \gamma\cdot\bm\omega_\textit{rand}.
\end{equation}
After the parameter inheriting, we keep training the selected model, to obtain its final optimal parameters:
\begin{equation}
\bm\omega^* = \mathop{\arg\min}_{\bm\omega} \mathcal{L}(\bm\psi^*,\bm\omega; \mathcal{D}^{t}).
\end{equation}

\section{Experiments}
In this section, we firstly describe the  experiment setting and implementation details (Section~\ref{sec:Experiment Setting and Implementation Details}). Then we show the results on zero-shot image classification (Section~\ref{sec:Zero-shot Image Classification}) and image-text retrieval (Section~\ref{sec:Zero-shot Image-Text Retrieval}). Finally, we conduct ablation study on our GrowCLIP to validate the effectiveness of components (Section~\ref{sec:Ablation Study and Sensitive Study}) and analyse (Section~\ref{sec:Analysis}).

\begin{table*}[t]
\small
\setlength{\tabcolsep}{6.5pt}
\renewcommand\arraystretch{0.92}

\begin{center}
\begin{tabular}{cccccccc cccc cc}
\toprule
\rotatebox{90}{\small{Growth step}}~~ &
\rotatebox{90}{\small{Method}}~~ &
\rotatebox{90}{\small{Training data}}~~ &
\rotatebox{90}{\small{Parameter}}~~ &
\rotatebox{90}{\small{Caltech101}}~~ &
\rotatebox{90}{\small{CIFAR10}}~~ &
\rotatebox{90}{\small{CIFAR100}}~~ &
\rotatebox{90}{\small{DTD}}~ &
\rotatebox{90}{\small{Flowers102}}~~ &
\rotatebox{90}{\small{Food101}}~~ &
\rotatebox{90}{\small{OxfordPets}}~~ &
\rotatebox{90}{\small{SUN397}}~~ &
\rotatebox{90}{\small{ImageNet}}~~ &
\rotatebox{90}{\small{\textbf{Average}}}~~ \\

\midrule
\multirow{6}{*}{Step 1}
& CLIP (TWP) & 3M &  149.6M & 42.8  & 37.6 & 14.5 & 10 & 9.3 & 16.1 & 21.6 & 22.0 & 14.0 & 20.9 \\
&CLIP (TFS)\cite{radford2021learning} & 3M &  149.6M & 42.8  & 37.6 & 14.5 & 10 & 9.3 & 16.1 & 21.6 & 22.0 & 14.0 & 20.9 \\
&MS-CLIP (TFS)\cite{you2022learning} & 3M & 129M & 47.1  & 35.5 & 14.4 & 10.0 & 12.9 & 17.3 & 30.1 & 24.7 & 18.1 & \textbf{23.3} \\
&CLIP (SAP)\cite{ash2020warm}  & 3M &  149.6M & 42.8  & 37.6 & 14.5 & 10 & 9.3 & 16.1 & 21.6 & 22.0 & 14.0 & 20.9 \\
&NAS-S1 & 3M & 30.0M & 40.5  & 31.2 & 13.3 & 9.0 & 8.9 & 15.8 & 20.2 & 23.2 & 13.7 & 19.5 \\
&GrowCLIP-S1 (ours) & 3M & 30.0M & 40.5  & 31.2 & 13.3 & 9.0 & 8.9 & 15.8 & 20.2 & 23.2 & 13.7 & 19.5 \\

\midrule
\multirow{6}{*}{Step 2}
&CLIP (TWP) & 6M &  149.6M & 50.3  & 36.1 & 16.0 & 11.3 & 12.4 & 21.2 & 27.1 & 25.8 & 17.9 & 24.2 \\
&CLIP (TFS)\cite{radford2021learning} & 6M & 149.6M & 58.5  & 52.0 & 22.3 & 12.1 & 13.1 & 25.0 & 33.6 & 30.1 & 20.9 & 29.7 \\
&MS-CLIP (TFS)\cite{you2022learning} & 6M & 129M & 60.6  & 50.0 & 18.3 & 13.4 & 17.4 & 27 & 42.4 & 32.6 & 25.2 & 31.8 \\
&CLIP (SAP)\cite{ash2020warm} & 6M & 149.6M & 61.9  & 49.0 & 21.4 & 12.4 & 15.5 & 24.6 & 35.4 & 31.6 & 22.6 & 30.5 \\
&NAS-S2 & 6M & 96.9M &  59.4 & 41.8 & 18.3 & 13.3 & 15.8 & 26.7 & 37.3 & 33.8 & 24.4 & 30.1 \\
&GrowCLIP-S2 (ours)& 6M & 116.6M & 57.7  & 48.0 & 21.2 & 13.7 & 18.0 & 29.5 & 39.5 & 34.5 & 25.7 & \textbf{32.0} \\

\midrule
\multirow{6}{*}{Step 3}
&CLIP (TWP) & 9M & 149.6M & 55.2  & 40.1 & 19.7 & 13.1 & 15.7 & 23.8 & 30.9 & 28.9 & 20.4 & 27.5 \\
&CLIP (TFS)\cite{radford2021learning} & 9M & 149.6M & 63.3  & 54.6 & 24.5 & 14.5 & 17.5 & 29.5 & 36.9 & 33.8 & 25.3 & 33.3 \\
&MS-CLIP (TFS)\cite{you2022learning} & 9M & 129M & 65.0  & 48.2 & 22.5 & 13.9 & 20.6 & 31.0 & 42.6 & 35.7 & 29.2 & 33.0 \\
&CLIP (SAP)\cite{ash2020warm} & 9M & 149.6M & 67.8  & 53.9 & 26.7 & 15.1 & 21.2 & 32.9 & 41.4 & 36.0 & 28.2 & 35.9 \\
&NAS-S3 & 9M & 287.4M & 66.7  & 46.8 & 22.2 & 14.9  & 19.7 & 35.4 & 46.1 & 39.3 & 30.1 & 35.8 \\
&GrowCLIP-S3 (ours)& 9M  & 168.6M  & 69.6  & 53.9 & 25.8 & 18.4 & 18.7 & 36.3 & 46.2 & 41.6 & 32.6 & \textbf{38.1} \\

\midrule
\multirow{6}{*}{Step 4}
&CLIP (TWP) & 12M &  149.6M & 57.5  & 44.4 & 20.5 & 12.1 & 14.2 & 24.1 & 34.7 & 30.3 & 22.2 & 28.9 \\
&CLIP (TFS)\cite{radford2021learning} & 12M & 149.6M & 65.9  & 57.3 & 29.7 & 16.3 & 18.5 & 32.9 & 45.5 & 36.7 & 28.3 & 36.8 \\
&MS-CLIP (TFS)\cite{you2022learning} & 12M & 129M & 65.3  & 50.9 & 26.3  & 16.2 & 21.9 & 35.9 & 50.1 & 38.4 & 32.0 & 37.4 \\
&CLIP (SAP)\cite{ash2020warm} & 12M & 149.6M & 69.8  & 60.6 & 31.9 & 12.4 & 21.0 & 36.7 & 47.0 & 40.4 & 32.2 & 39.1 \\
&NAS-S4 & 12M & 325.9M & 70.8  & 53.9 & 28.3 & 16.7 & 21.7 & 39.5 & 51.2 & 41.3 & 34.2 & 39.7 \\
&GrowCLIP-S4 (ours)& 12M & 188.4M & 71.9  & 60.7 & 28.3 & 17.3 & 23.3 & 42.5 & 52.4 & 45.5 & 36.1 & \textbf{42.0} \\

\end{tabular}
\end{center}
\caption{Top-1 accuracy(\%) of zero-shot image classification on 9 datasets. Our GrowCLIP outperforms the baseline methods in terms of average top-1 accuracy over $9$ datasets at each growth step except step $1$.}
\label{zeroshot-classification-table}
\end{table*}

\begin{table*}[t]
\begin{center}

\setlength{\tabcolsep}{5pt}
\renewcommand\arraystretch{0.88}
\small
\begin{tabular}{cccccccccccccc}
\toprule
    \multirow{3}{*}{Growth step}    &  \multirow{3}{*}{Method}  & \multicolumn{6}{c}{Flickr30K}                                                               & \multicolumn{6}{c}{MSCOCO}                                                                  \\
      &      & \multicolumn{3}{c}{image-to-text} & \multicolumn{3}{c}{text-to-image} & \multicolumn{3}{c}{image-to-text} & \multicolumn{3}{c}{text-to-image} \\
      &      & R@1           & R@5           & R@10         & R@1           & R@5           & R@10         & R@1           & R@5           & R@10         & R@1           & R@5           & R@10         \\
\midrule
\multirow{6}{*}{Step 1}
&CLIP (TWP) & 19.6          & 45.3          & 56.4           & 14.3          & 33.5        & 44.1         & 10.7            & 26.3          & 36.8         & 7.0          & 19.1            & 27.5         \\
&CLIP (TFS) \cite{radford2021learning} & 19.6          & 45.3          & 56.4           & 14.3          & 33.5        & 44.1         & 10.7            & 26.3          & 36.8         & 7.0          & 19.1            & 27.5         \\
&MS-CLIP (TFS)\cite{you2022learning} & \textbf{27.4}          & \textbf{51.8}           & \textbf{63.3}            & \textbf{18.1}           & \textbf{39.2}           & \textbf{50.7}          & \textbf{12.9}             & \textbf{31.9}           & \textbf{43.2}          & \textbf{8.8}           & \textbf{23.5}             & \textbf{33.1}          \\
&CLIP (SAP) \cite{ash2020warm} & 19.6          & 45.3          & 56.4           & 14.3          & 33.5        & 44.1         & 10.7            & 26.3          & 36.8         & 7.0          & 19.1            & 27.5         \\
&NAS-S1  & 22.8          & 45.2          & 57.8           & 13.5          & 34.2          & 46.0         & 10.6            & 27.4          & 37.5         & 7.0          & 19.9            & 28.8         \\
&GrowCLIP-S1 (ours)   & 22.8          & 45.2          & 57.8           & 13.5          & 34.2          & 46.0         & 10.6            & 27.4          & 37.5         & 7.0          & 19.9            & 28.8         \\
\midrule
\multirow{6}{*}{Step 2}
&CLIP (TWP) & 26.6          & 51.0          & 62.0           & 17.4          & 39.8          & 50.7         & 12.8            & 30.5          & 40.8         & 8.3          & 22.5            & 31.9         \\
&CLIP (TFS) \cite{radford2021learning} & 31.4          & 57.7          & 68.1           & 21.2          & 46.7          & 57.1         & 17.1            & 37.4          & 48.7         & 10.9          & 27.3            & 37.4         \\
&MS-CLIP (TFS)\cite{you2022learning} & 35.8          & \textbf{65.1}          & \textbf{75.0}           & 25.8          & \textbf{53.1}          & \textbf{64.6}         & \textbf{19.3}            & \textbf{43.0}          & \textbf{55.2}         & \textbf{13.1}          & \textbf{31.5}            & \textbf{42.6}         \\
&CLIP (SAP) \cite{ash2020warm}      & 33.6          & 60.9          & 72.0           & 24.1          & 48.2          & 59.3         & 17.1            & 38.0          & 50.2         & 10.9          & 28.0            & 28.1         \\
&NAS-S2  & 36.9          & 63.0          & 74.0           & 25.2          & 51.7          & 62.3         & 18.4            & 41.6          & 53.7         & 12.9          & 30.9            & 41.7         \\
&GrowCLIP-S2  (ours)      & \textbf{37.5}            & 63.7          & 74.2         & \textbf{25.9}          & 51.1          & 62.8         & 19.1          & 42.8          & 54.7         & 12.9         & 30.8          & 41.9         \\
\midrule
\multirow{6}{*}{Step 3}
&CLIP (TWP) & 28.3          & 54.4          & 66.6           & 20.6          & 42.2          & 54.7         & 14.4            & 34.1          & 45.0         & 9.6          & 25.2            & 34.9         \\
&CLIP (TFS) \cite{radford2021learning} & 37.5          & 63.5          & 74.2           & 26.0          & 52.1          & 64.0         & 19.7            & 43.2          & 55.1         & 13.3          & 31.8            & 42.6         \\
&MS-CLIP (TFS)\cite{you2022learning} & 40.0          & 67.7          & 78.2           & 29.8          & 57.0          & 68.8         & 22.9            & 47.9          & 59.1         & 15.3          & 35.2            & 46.6         \\
&CLIP (SAP) \cite{ash2020warm}      & 40.7          & 68.0          & 78.3           & 28.1          & 55.3          & 67.1         & 22.8            & 45.7          & 58.8         & 14.2          & 33.4            & 44.6         \\
&NAS-S3  & 43.2          & 72.8          & \textbf{80.7}           & 32.1          & 59.9          & 70.2         & 23.4            & 48.3          & \textbf{61.0}         & 15.7          & 36.8           & 48.3         \\
&GrowCLIP-S3  (ours)   & \textbf{44.6}          & \textbf{73.1}          & \textbf{80.7}           & \textbf{33.2}          & \textbf{60.7}          & \textbf{70.9}         & \textbf{23.5}            & \textbf{48.8}        & \textbf{61.0}           & \textbf{15.8}          & \textbf{36.9}            & \textbf{48.4}         \\
\midrule
\multirow{6}{*}{Step 4}
&CLIP (TWP) & 30.2          & 56.3          & 66.3           & 21.7          & 46.6          & 57.6         & 15.6            & 36.2          & 47.8         & 10.4          & 26.5            & 36.9         \\
&CLIP (TFS) \cite{radford2021learning} & 41.5          & 68.6          & 80.0           & 30.0          & 56.9          & 68.0         & 21.8            & 45.5          & 57.3         & 14.3          & 33.8            & 45.1         \\
&MS-CLIP (TFS)\cite{you2022learning} & 44.3          & 72.8          & 81.1           & 32.6          & 61.3          & 72.1         & 25.2            & 50.5          & 62.0         & 16.6          & 37.7            & 49.0         \\
&CLIP (SAP) \cite{ash2020warm}      & 46.9          & 74.7          & 83.7           & 33.7          & 61.3          & 72.1         & 25.2            & 50.8          & 62.8         & 16.6          & 37.4            & 49.1         \\
&NAS-S4  &    49.6       &     75.2      &     83.3       &      37.0     &      65.2     &     75.4     &       26.8      &      53.0     &      64.6    &     17.7      &     39.6        &    51.5      \\
&GrowCLIP-S4  (ours)    & \textbf{50.8}          & \textbf{75.8}          & \textbf{84.0}           & \textbf{37.2}          & \textbf{65.3}          & \textbf{76.2}         & \textbf{27.6}            & \textbf{53.2}          & \textbf{64.9}         & \textbf{18.1}          & \textbf{40.3}            & \textbf{52.0}         \\
\bottomrule         
\end{tabular}
\end{center}
\caption{Results of zero-shot image-text retrieval on Flickr30K and MSCOCO datasets. R@K means top-K recall.}
\label{tab:zero-shot-retrieval-table}
\end{table*}

\subsection{Experiment Setting}\label{sec:Experiment Setting and Implementation Details}

\textbf{Benchmark Protocol}  \;Conceptual 12M (CC12M) \cite{changpinyo2021cc12m} collects about $12$ million image-text pairs from  the Internet. The CC12M is randomly divided into $4$ subsets. We select one subset as the training set for the first growth step and adds one subset to the training set at each subsequent growth step such that there are total $4$ growth steps in our experiments. At each growth step, we evaluate the corresponding model on the downstream tasks including zero-shot image classification and zero-shot image-text retrieval. The grown architecture of each growth step is shown in Appendix. 

\textbf{Comparison Methods}  \;To demonstrate the effectiveness of our method, we compare it with the following baselines: (i) Training with pre-training (TWP) is a usual method in transfer learning. (ii) Training from scratch (TFS) \cite{radford2021learning} is used as a basic method to train language-image pre-trained model. (iii) MS-CLIP \cite{you2022learning} is a hand-crafted framework with a shareable module and modality-specific auxiliary module (the early specialization). (iv) The shrink and perturb (SAP) \cite{ash2020warm} is a method of parameter inheriting in warm-starting learning. (v) Neural architecture search (NAS) \cite{peng2020cream,chen2021autoformer} is an important research field of auto machine learning, in which we use the same search space as the growth space of GrowCLIP. 

\textbf{Implementation Details} \;\label{Implementation Details} A dual-stream model following CLIP \cite{radford2021learning} without shared encoder is used as basic architecture at the first growth step. The image encoder is stacked with $6$ transformer blocks with the following hyperparameters: number of self-attention heads is $6$ and the hidden layer size is $384$. As for the text encoder, $6$ transformer blocks are stacked and the number of the attention heads and the width are $4$ and $256$ respectively. The resolution of input in the images encoder is resized to $224 \time 224$ and the maximum sequence in the text encoder is limited to $77$.
The center crop is the only data augmentation method for training. We use the prompts like CLIP\footnote{\url{https://github.com/openai/CLIP/blob/main/data/prompts.md}} on the zero-shot image classification. We adopt LAMB optimizer \cite{2019Large} with learning rate $0.01$ and weight decay $0.0001$, and cosine learning rate schedule \cite{2016SGDR} with a $4000$-iteration linear warmup. For the hyperparameters, the proportion of the old model and the new one in the parameter inheriting  
$\beta = 0.3$, $\gamma = 0.001$. The model parameter coefficient in growth architecture selection $\alpha = 0.5$. We train the model with a total batch size of $1536$ for $30$ epochs at each growth step, which includes $2$ epochs for supernet fine-tuning, $2$ epochs for supernet training and $26$ epochs for grown model training. For the baselines, the CLIP adopts the same usual architecture (ViT-B/16), which has 12-layer 768-wide transformer blocks with 12 attention heads for image encoder and 12-layer 512-wide transformer blocks with 8 attention heads for text encoder and are trained for 30 epochs. For NAS, we first train the supernet for 30 epochs to select the best architecture, and then train the selected model from scratch for another 30 epochs. All experiments are performed on 16 NVIDIA Tesla V100 GPUs.

\subsection{Zero-shot Image Classification}\label{sec:Zero-shot Image Classification}
At each growth step, we evaluate the model on zero-shot image classification task. The downstream datasets includes $9$ typical settings, Caltech101 \cite{fei2006one}, CIFAR10 \cite{krizhevsky2009learning}, CIFAR100 \cite{krizhevsky2009learning}, Describable Textures Dataset (DTD) \cite{cimpoi2014describing}, Flowers102 \cite{nilsback2008automated}, Food101 \cite{bossard2014food}, OxfordPets \cite{parkhi2012cats}, SUN39 \cite{barriuso2012notes} and ImageNet \cite{deng2009large}. We apply prompt ensemble evaluation like CLIP for each dataset. Results are given in Table~\ref{zeroshot-classification-table}. Our model performs growth process at each growth step except step $1$, where the parameter grows from $30.0M$ to $188.4M$. It outperforms the baseline methods in terms of average top-1 accuracy over $9$ datasets at the last three growth steps. For example, at the third growth step, our model improves the average top-1 accuracy from $35.9\%$ to $38.1\%$ ($+2.2\%$).

\subsection{Zero-shot Image-Text Retrieval}\label{sec:Zero-shot Image-Text Retrieval}
The zero-shot image-text retrieval consists of two sub-tasks: image-to-text retrieval and text-to-image retrieval. We evaluate the model of each growth step on two retrieval benchmark datasets: Flickr30K \cite{plummer2015flickr30k} and MSCOCO \cite{lin2014microsoft}. As shown in Table~\ref{tab:zero-shot-retrieval-table}, our model gets better as model grows and data grows. At the last two growth steps, it can be observed that our method has great performance on all metrics. Particularly, our method improves from $49.6\%$ to $50.8\%$ ($+1.2\%$) for top-$1$ image-to-text recall on Flickr30K dataset and from $26.8\%$ to $27.6\%$ ($+0.8\%$) on MSCOCO at the last growth step.

\subsection{Ablation Study}\label{sec:Ablation Study and Sensitive Study}
Here, we conduct the ablation study on our GrowCLIP. All experiments are built at the growth step 2 and we evaluate the models on zero-shot image classification of ImageNet. To verify the effectiveness of each module in our proposed GrowCLIP, we conduct ablation experiments at some core steps mentioned in Section~\ref{sec:pipeline} as follows. (i) Parameter inheriting with momentum in phase $1$ (PIM@1). (ii) Supernet fine-tuning in phase $1$ (ST). (iii) Growth architecture selection in phase $2$ (GAS). (iv) Parameter inheriting with momentum in phase $3$ (PIM@3). The accuracy on zero-shot image classification of ImageNet is reported for comparison and the results of the ablation study are shown in Table~\ref{tab:ablation}. As can be seen, all above four components have positive effects on the final results. Specifically, the supernet training will influence the grown model without PIM@3, which results in $1.2\%$ drop. Compared with GAS method, we randomly choose an architecture in the growth space, which has a worse performance from $24.5\%$ to $22.6\%$. With ST, we observe a remarkable improvement of $1.5\%$. GrowCLIP with PIM@1 solves the problem of local minimum dilemma and brings a $0.3\%$ improvement. 

\begin{table}
\begin{center}
\setlength{\tabcolsep}{5pt}{
\begin{tabular}{cccccc}
    \toprule
    PIM@1 &
    ST &
    GAS  & 
    PIM@3 &
    Para. &
    Acc. \\
    \midrule
    \cmark & \cmark & \cmark &  \cmark & 116.6M & \textbf{25.7} \\
    \cmark & \cmark & \cmark &  \xmark & 116.6M & 24.5  \\
    \cmark & \cmark & \xmark &  \xmark & 109.5M & 22.6  \\
    \cmark & \xmark & \xmark &  \xmark & 109.5M & 21.1  \\
    \xmark & \xmark & \xmark &  \xmark & 109.5M & 20.8  \\
    \bottomrule
\end{tabular}}
\end{center}
\caption{Ablation analysis of each module in GrowCLIP at Growth Step 2. ($\text{Para.}$: parameter of the model. $\text{Acc.}$: top-1 accuracy (\%) of zero-shot image classification on ImageNet.)}
\label{tab:ablation}
\end{table}

\subsection{Analysis}\label{sec:Analysis}

\textbf{Why can GrowCLIP improve the performance?} \; Compared with baseline models, the gains of our GrowCLIP come from two folds: (i) GrowCLIP can search the architecture that matches the current data with Growth architecture selection (GAS). (ii) Parameter inheriting with momentum (PIM) benefits GrowCLIP from yesterday's model. To demonstrate their effectiveness, we train the architecture selected by GrowCLIP at step $4$ from scratch, and the result is shown in Table~\ref{tab:Ablation analysis of GrowCLIP-S4}. Compared with ViT-B/16, the growth architecture selection can bring $5\%$ improvement on  zero-shot image classification of ImageNet. Furthermore, it is even better than ViT-L/16, which is two and a half times as big as our model. More experiments have proved the effectiveness of GAS in the appendix. 
Comparing our GrowCLIP-S4 with GrowCLIP-S4 (TFS), we observe a remarkable improvement of $2.8\%$ with PIM.

\begin{table}[t]
\begin{center}
\setlength{\tabcolsep}{15pt}
\begin{tabular}{lcc}
\toprule
 &
Para. &
Acc. \\
\midrule
CLIP-ViT-B/16 (TFS)      & 149.6M & 28.3     \\
CLIP-ViT-L/16 (TFS)      & 431.1M & 29.9   \\
GrowCLIP-S4 (TFS)   & 188.4M & 33.3    \\
GrowCLIP-S4 (ours)  & 188.4M & \textbf{36.1}    \\
\bottomrule
\end{tabular}
\end{center}
\caption{Ablation analysis of GrowCLIP-S4. }
\label{tab:Ablation analysis of GrowCLIP-S4}
\end{table}

\textbf{Why can GrowCLIP improve the training efficiency?} \; Our GrowCLIP can improve training efficiency, due to GAS and PIM, and the result is shown in Table~\ref{tab:efficience analysis}.
(i) Compared with the fixed architecture method, GrowCLIP has a smaller model with the GAS, so we reduce training costs at the early steps (48 GPU hours saved at step 2). At the later steps, GrowCLIP uses fewer epochs to achieve good performance because of PIM. As shown in Figure~\ref{fig:1}, GrowCLIP has far outperformed CLIP (TFS) only after supernet training. To demonstrate the effectiveness of PIM, we compare GrowCLIP-S4 with GrowCLIP-S4 (TFS). GrowCLIP only uses 9 epochs to achieve comparable performance with PIM.
(ii) Compared with the NAS method,  GrowCLIP inherits the parameters of the old model with momentum, which only takes 2 epochs to fine-tune the supernet and 2 epochs to train the supernet. In contrast, one-shot NAS methods consume about $3.3$X resources.

\begin{table}[t]
\begin{center}
\setlength{\tabcolsep}{5pt}
\begin{tabular}{lcccc}
\toprule
 &
Para. &
Acc. &
Epoch &
Cost
\\
\midrule
CLIP-S2(TFS)  & 149.6M &  20.9 & 30 & 336  \\
GrowCLIP-S2 (ours) & 116.6M &  25.7 & 30 & 288     \\
\midrule
GrowCLIP-S4 (TFS)  & 188.4M &  33.3 & 30 & 1230  \\
GrowCLIP-S4 (ours) & 188.4M &  33.3 & 9 & 722     \\
\midrule
NAS-S4  & 325.9M  &  34.2 & 60 & 5280   \\
GrowCLIP-S4 (ours) & 188.4M &  36.1 & 30 & 1583     \\
\bottomrule
\end{tabular}
\end{center}
\caption{Efficiency analysis. (Cost: GPU hours)}
\label{tab:efficience analysis}
\end{table}

\textbf{Can GrowCLIP find the relationship between model and data?} \; When the model scale is sufficient large, there is no need to enlarge the model. We conduct the experiment started from ViT-L/16. And the result in Table~\ref{bigger initialized} shows that the larger model startpoint hardly grows at step 2, which means that our GrowCLIP can find the relationship between the model and data. In our online learning scenario, the initial architecture of the model is not important due to our growth pipeline. If the model scale is smaller at the beginning, GrowCLIP will select a larger enough model for growth. In contrast, if the model scale is too large initially, GrowCLIP will stop growing.

\begin{table}[t]
\begin{center}
\setlength{\tabcolsep}{20pt}
\begin{tabular}{lcc}
\toprule
 &
Para. &
Acc. \\
\midrule
GrowCLIP-S1   & 431.1M & 14.6     \\
GrowCLIP-S2  & 431.2M & 26.2    \\
\bottomrule
\end{tabular}
\end{center}
\caption{The architecture of GrowCLIP-S1 with bigger initialized model (ViT-L/16). }
\label{bigger initialized}
\end{table}

\begin{table}[t]
\begin{center}
\setlength{\tabcolsep}{1.8pt}
\begin{tabular}{ccccccc}
\toprule
 \multicolumn{1}{c}{}  & \multicolumn{2}{c}{Step 2} 
  & \multicolumn{2}{c}{Step 3} 
   & \multicolumn{2}{c}{Step 4} \\
\midrule
Dimension &
Para. &
Acc. &
Para. &
Acc. &
Para. &
Acc. \\
\midrule
+ $4$ $b^I$ & +7.1M & +0.93 & +19.7M & +0.46 & +19.7M & +0.12 \\
+ $4$ $h^I$& +20.1M & +0.34 &  +67.3M & +0.12  &  +105.1M & +0.03    \\
+ $2$ $l^I$& +0.1M & +0.84  &  +0.1M & +0.07  &  +0.1M & +0.10  \\
+ $4$ $b^T$ & +3.2M & +0.01  &  +12.6M & +0.08  &  +12.6M & +0.01    \\
+ $4$ $h^T$& +27.0M & +0.22  &  +45.0M & +0.01  &  +69.5M & -0.04   \\
+ $4$ $b^S$ & +7.1M & +1.03  &  +19.7M & +0.19  &  +19.7M & -0.07    \\
\bottomrule
\end{tabular}
\end{center}
\caption{The ablation of growth space corresponding to different dimensions. (+ $4$ $b^I$: the number of transformer blocks in image encoder is increased by $4$)}
\label{tab:Ablation analysis of different components}
\end{table}

\textbf{What is the influence when growth happens in different dimensions?} \; As shown in Table~\ref{tab:Ablation analysis of different components}, different growth dimensions have different influences on the performance. By analyzing the experimental results, we conclude $4$ laws, which can provide a reference for the design of cross-modal model architecture:
(i) Enlarging the smaller model bring a greater improvement than the bigger one. For example, when the growth appears in transformer blocks in the image encoder, the accuracy can gain $0.93\%$ improvement at step 2 but only $0.12\%$ at step 4.
(ii) We need to leave the image encoder with more capacity. Compared with text encoder, the improvement of growing the image encoder is more significant at each step.
(iii) The convolutional layers is a very important component in the image encoder, which can help the model to better extract visual feature.
(iv) The shared encoder can enhance the degree of cross-modal fusion and bring a remarkable improvement. However, do not use a large number of transformers in the shared encoder, or the model become redundant with no noticeable performance improvement.

\section{Conclusion}
\label{sec:Conclusion}
In this paper, we study the online learning case
in cross-modal pre-trained models. The small-scale architecture may limit the final performance due to the limited parameters when data grows. Based on this real-world scenario, we propose a data-driven automatic model growing algorithm for language-image pre-training - GrowCLIP, which upgrades the model size when abundant new data arrives. We adopt parameter inheriting with momentum (PIM) to utilize the knowledge in yesterday's model and seek out the optimal growth architecture at each growth step with growth architecture selection (GAS).
GrowCLIP improves the performance on downstream tasks, including zero-shot image classification and retrieval.
The effectiveness of each component of GrowCLIP is proved on ablation studies. 
And we provide a reference for the design of cross-modal model architecture.

\textbf{Limitations} \;  
In this work, we only prove the effectiveness of GrowCLIP with public pre-training dataset - CC12M.
In future works, we will extend our method to a real-world scenario, where the VLP model keeps training with constantly updated data crawled from the web.

\section*{Acknowledgment}

This work was supported in part by National Key R$\&$D Program of China under
Grant No.2020AAA0109700, Guangdong Outstanding Youth Fund (Grant No. 2021B1515020061), Shenzhen Science and Technology Program (Grant No.RCYX20200714114642083), Shenzhen Fundamental Research Program(Grant No. JCYJ20190807154211365), Nansha Key RD Program under Grant No.2022ZD014 and Sun Yat-sen University under Grant No.22lgqb38 and 76160-12220011. We thank MindSpore for the partial support of this work, which is a new deep learning computing framework\footnote{https://www.mindspore.cn/}. 

\newpage

{\small
\bibliographystyle{ieee_fullname}
\bibliography{egbib}

\begin{thebibliography}{10}\itemsep=-1pt

\bibitem{alayracflamingo}
J. Alayrac, J. Donahue, P. Luc, A. Miech, I. Barr, Y. Hasson, K. Lenc, A.
  Mensch, K. Millican, M. Reynolds, et~al.
\newblock Flamingo: a visual language model for few-shot learning.
\newblock In {\em Advances in Neural Information Processing Systems}.

\bibitem{aljundi2018memory}
R. Aljundi, F. Babiloni, M. Elhoseiny, M. Rohrbach, and T. Tuytelaars.
\newblock Memory aware synapses: Learning what (not) to forget.
\newblock In {\em European conference on computer vision}, 2018.

\bibitem{antol2015vqa}
S. Antol, A. Agrawal, J. Lu, M. Mitchell, D. Batra, C. Zitnick, and D. Parikh.
\newblock Vqa: Visual question answering.
\newblock In {\em International Conference on Computer Vision}, 2015.

\bibitem{ash2020warm}
J. Ash and R. Adams.
\newblock On warm-starting neural network training.
\newblock 2020.

\bibitem{barriuso2012notes}
A. Barriuso and A. Torralba.
\newblock Notes on image annotation.
\newblock Preprint arXiv:1210.3448, 2012.

\bibitem{bossard2014food}
L. Bossard, M. Guillaumin, and L. Gool.
\newblock Food-101--mining discriminative components with random forests.
\newblock In {\em European Conference on Computer Vision}, 2014.

\bibitem{buzzega2020dark}
P. Buzzega, M. Boschini, A. Porrello, D. Abati, and S. Calderara.
\newblock Dark experience for general continual learning: a strong, simple
  baseline.
\newblock In {\em Advances in Neural Information Processing Systems}, 2020.

\bibitem{cai2019once}
H. Cai, C. Gan, T. Wang, Z. Zhang, and S. Han.
\newblock Once-for-all: Train one network and specialize it for efficient
  deployment.
\newblock In {\em International Conference on Learning Representations}, 2019.

\bibitem{changpinyo2021cc12m}
S. Changpinyo, P. Sharma, N. Ding, and R. Soricut.
\newblock {Conceptual 12M}: Pushing web-scale image-text pre-training to
  recognize long-tail visual concepts.
\newblock In {\em Computer Vision and Pattern Recognition}, 2021.

\bibitem{chen2021autoformer}
M. Chen, H. Peng, J. Fu, and H. Ling.
\newblock Autoformer: Searching transformers for visual recognition.
\newblock In {\em International Conference on Computer Vision}, 2021.

\bibitem{chen2020simple}
T. Chen, S. Kornblith, M. Norouzi, and G. Hinton.
\newblock A simple framework for contrastive learning of visual
  representations.
\newblock In {\em International Conference on Machine Learning}, 2020.

\bibitem{cimpoi2014describing}
M. Cimpoi, S. Maji, I. Kokkinos, S. Mohamed, and A. Vedaldi.
\newblock Describing textures in the wild.
\newblock In {\em Computer Vision and Pattern Recognition}, 2014.

\bibitem{cornia2019show}
M. Cornia, L. Baraldi, and R. Cucchiara.
\newblock Show, control and tell: A framework for generating controllable and
  grounded captions.
\newblock In {\em Computer Vision and Pattern Recognition}, 2019.

\bibitem{deng2009large}
J. Deng.
\newblock A large-scale hierarchical image database.
\newblock 2009.

\bibitem{devlin2019bert}
J. Devlin, M. Chang, K. Lee, and K. Toutanova.
\newblock {BERT: Pre-training of Deep Bidirectional Transformers for Language
  Understanding}.
\newblock In {\em North American Chapter of the Association for Computational
  Linguistics}, 2019.

\bibitem{dosovitskiy2021image}
A. Dosovitskiy, L. Beyer, A. Kolesnikov, D. Weissenborn, X. Zhai, T.
  Unterthiner, M. Dehghani, M. Minderer, G. Heigold, S. Gelly, et~al.
\newblock An image is worth 16x16 words: Transformers for image recognition at
  scale.
\newblock In {\em International Conference on Learning Representations}, 2021.

\bibitem{grabner2006line}
H. Grabner and H. Bischof.
\newblock On-line boosting and vision.
\newblock In {\em Computer Vision and Pattern Recognition}, 2006.

\bibitem{grill2020bootstrap}
J. Grill, F. Strub, F. Altch{\'e}, C. Tallec, P. Richemond, E. Buchatskaya, C.
  Doersch, B. Avila~Pires, Z. Guo, M. Gheshlaghi~Azar, et~al.
\newblock Bootstrap your own latent-a new approach to self-supervised learning.
\newblock 2020.

\bibitem{guo2020bootstrap}
Z. Guo, B.~Avila Pires, B. Piot, J. Grill, F. Altch{\'e}, R. Munos, and M.
  Azar.
\newblock Bootstrap latent-predictive representations for multitask
  reinforcement learning.
\newblock In {\em International Conference on Machine Learning}, 2020.

\bibitem{guo2020single}
Z. Guo, X. Zhang, H. Mu, W. Heng, Z. Liu, Y. Wei, and J. Sun.
\newblock Single path one-shot neural architecture search with uniform
  sampling.
\newblock In {\em European Conference on Computer Vision}, 2020.

\bibitem{hadsell2020embracing}
R. Hadsell, D. Rao, A. Rusu, and R. Pascanu.
\newblock Embracing change: Continual learning in deep neural networks.
\newblock {\em Trends in Cognitive Sciences}, 2020.

\bibitem{hazan2016introduction}
E. Hazan et~al.
\newblock Introduction to online convex optimization.
\newblock {\em Foundations and Trends{\textregistered} in Optimization}, 2016.

\bibitem{he2021masked}
K. He, X. Chen, S. Xie, Y. Li, P. Doll{\'a}r, and R. Girshick.
\newblock Masked autoencoders are scalable vision learners.
\newblock In {\em Conference on Computer Vision and Pattern Recognition}, 2022.

\bibitem{he2020momentum}
K. He, H. Fan, Y. Wu, S. Xie, and R. Girshick.
\newblock Momentum contrast for unsupervised visual representation learning.
\newblock In {\em Computer Vision and Pattern Recognition}, 2020.

\bibitem{hu2021scaling}
X. Hu, Z. Gan, J. Wang, Z. Yang, Z. Liu, Y. Lu, and L. Wang.
\newblock Scaling up vision-language pre-training for image captioning.
\newblock Preprint arXiv:2111.12233, 2021.

\bibitem{hu2020vivo}
X. Hu, X. Yin, K. Lin, L. Wang, L. Zhang, J. Gao, and Z. Liu.
\newblock Vivo: Surpassing human performance in novel object captioning with
  visual vocabulary pre-training.
\newblock 2020.

\bibitem{jia2021scaling}
C. Jia, Y. Yang, Y. Xia, Y. Chen, Z. Parekh, H. Pham, Q. Le, Y. Sung, Z. Li,
  and T. Duerig.
\newblock Scaling up visual and vision-language representation learning with
  noisy text supervision.
\newblock In {\em International Conference on Machine Learning}, 2021.

\bibitem{kaplan2020scaling}
J. Kaplan, S. McCandlish, T. Henighan, T. Brown, B. Chess, R. Child, S. Gray,
  A. Radford, J. Wu, and D. Amodei.
\newblock Scaling laws for neural language models.
\newblock Preprint arXiv:2001.08361, 2020.

\bibitem{kim2021vilt}
W. Kim, B. Son, and I. Kim.
\newblock Vilt: Vision-and-language transformer without convolution or region
  supervision.
\newblock In {\em International Conference on Machine Learning}, 2021.

\bibitem{kirkpatrick2017overcoming}
J. Kirkpatrick, R. Pascanu, N. Rabinowitz, J. Veness, G. Desjardins, A. Rusu,
  K. Milan, J. Quan, T. Ramalho, A. Grabska-Barwinska, et~al.
\newblock Overcoming catastrophic forgetting in neural networks.
\newblock {\em National Academy of Sciences}, 2017.

\bibitem{krizhevsky2009learning}
A. Krizhevsky, G. Hinton, et~al.
\newblock Learning multiple layers of features from tiny images.
\newblock 2009.

\bibitem{lan2019albert}
Z. Lan, M. Chen, S. Goodman, K. Gimpel, P. Sharma, and R. Soricut.
\newblock {ALBERT: A Lite BERT for Self-supervised Learning of Language
  Representations}.
\newblock In {\em International Conference on Learning Representations}, 2020.

\bibitem{li2022automated}
C Li, B. Zhuang, G. Wang, X. Liang, X. Chang, and Y. Yang.
\newblock Automated progressive learning for efficient training of vision
  transformers.
\newblock In {\em Computer Vision and Pattern Recognition}, 2022.

\bibitem{fei2006one}
F. Li, R. Fergus, and P. Perona.
\newblock One-shot learning of object categories.
\newblock {\em IEEE transactions on pattern analysis and machine intelligence},
  2006.

\bibitem{li2022blip}
J. Li, D. Li, C. Xiong, and S. Hoi.
\newblock Blip: Bootstrapping language-image pre-training for unified
  vision-language understanding and generation.
\newblock In {\em International Conference on Machine Learning}, pages
  12888--12900. PMLR, 2022.

\bibitem{li2019visualbert}
L. Li, M. Yatskar, D. Yin, C. Hsieh, and K. Chang.
\newblock Visualbert: A simple and performant baseline for vision and language.
\newblock Preprint arXiv:1908.03557, 2019.

\bibitem{li2020oscar}
X. Li, X. Yin, C. Li, P. Zhang, X. Hu, L. Zhang, L. Wang, H. Hu, L. Dong, F.
  Wei, et~al.
\newblock Oscar: Object-semantics aligned pre-training for vision-language
  tasks.
\newblock In {\em European Conference on Computer Vision}, 2020.

\bibitem{li2021supervision}
Y. Li, F. Liang, L. Zhao, Y. Cui, W. Ouyang, J. Shao, F. Yu, and J. Yan.
\newblock Supervision exists everywhere: A data efficient contrastive
  language-image pre-training paradigm.
\newblock Preprint arXiv:2110.05208, 2021.

\bibitem{lin2014microsoft}
T. Lin, M. Maire, S. Belongie, J. Hays, P. Perona, D. Ramanan, P. Doll{\'a}r,
  and C. Zitnick.
\newblock Microsoft coco: Common objects in context.
\newblock In {\em European Conference on Computer Vision}, 2014.

\bibitem{liu2018progressive}
C. Liu, B. Zoph, M. Neumann, J. Shlens, W. Hua, L. Li, F. Li, A. Yuille, J.
  Huang, and K. Murphy.
\newblock Progressive neural architecture search.
\newblock In {\em European conference on computer vision}, 2018.

\bibitem{liu2018darts}
Hanxiao Liu, Karen Simonyan, and Yiming Yang.
\newblock Darts: Differentiable architecture search.
\newblock Preprint arXiv:1806.09055, 2018.

\bibitem{liu2021cptr}
W. Liu, S. Chen, L. Guo, X. Zhu, and J. Liu.
\newblock Cptr: Full transformer network for image captioning.
\newblock Preprint arXiv:2101.10804, 2021.

\bibitem{liu2019roberta}
Y. Liu, M. Ott, N. Goyal, J. Du, M. Joshi, D. Chen, O. Levy, M. Lewis, L.
  Zettlemoyer, and V. Stoyanov.
\newblock {RoBERTa: A Robustly Optimized BERT Pretraining Approach}.
\newblock Preprint arXiv:1907.11692, 2019.

\bibitem{2016SGDR}
I. Loshchilov and F. Hutter.
\newblock Sgdr: Stochastic gradient descent with warm restarts.
\newblock 2016.

\bibitem{lu202012}
J. Lu, V. Goswami, M. Rohrbach, D. Parikh, and S. Lee.
\newblock 12-in-1: Multi-task vision and language representation learning.
\newblock In {\em Computer Vision and Pattern Recognition}, 2020.

\bibitem{nilsback2008automated}
M. Nilsback and A. Zisserman.
\newblock Automated flower classification over a large number of classes.
\newblock In {\em Computer Vision, Graphics \& Image Processing}, 2008.

\bibitem{parkhi2012cats}
O. Parkhi, A. Vedaldi, A. Zisserman, and C. Jawahar.
\newblock Cats and dogs.
\newblock In {\em Computer Vision and Pattern Recognition}, 2012.

\bibitem{peng2020cream}
H. Peng, H. Du, H. Yu, Q. Li, J. Liao, and J. Fu.
\newblock Cream of the crop: Distilling prioritized paths for one-shot neural
  architecture search.
\newblock In {\em Advances in Neural Information Processing Systems}, 2020.

\bibitem{plummer2015flickr30k}
B. Plummer, L. Wang, C. Cervantes, J. Caicedo, J. Hockenmaier, and S. Lazebnik.
\newblock Flickr30k entities: Collecting region-to-phrase correspondences for
  richer image-to-sentence models.
\newblock In {\em International Conference on Computer Vision}, 2015.

\bibitem{pogodin2020first}
R. Pogodin and T. Lattimore.
\newblock On first-order bounds, variance and gap-dependent bounds for
  adversarial bandits.
\newblock In {\em Uncertainty in Artificial Intelligence}, 2020.

\bibitem{radford2021learning}
A. Radford, J. Kim, C. Hallacy, A. Ramesh, G. Goh, S. Agarwal, G. Sastry, A.
  Askell, P. Mishkin, J. Clark, et~al.
\newblock Learning transferable visual models from natural language
  supervision.
\newblock In {\em International Conference on Machine Learning}, 2021.

\bibitem{real2018regularized}
E. Real, A. Aggarwal, Y. Huang, and Q. Le.
\newblock {Regularized Evolution for Image Classifier Architecture Search}.
\newblock In {\em AAAI Conference on Artificial Intelligence}, 2019.

\bibitem{real2017large}
E. Real, S. Moore, A. Selle, S. Saxena, Y. Suematsu, J. Tan, Q. Le, and A.
  Kurakin.
\newblock Large-scale evolution of image classifiers.
\newblock In {\em International Conference on Machine Learning}, 2017.

\bibitem{saffari2009line}
A. Saffari, C. Leistner, J. Santner, M. Godec, and H. Bischof.
\newblock On-line random forests.
\newblock In {\em International Conference on Computer Vision workshops}, 2009.

\bibitem{shalev2012online}
S. Shalev-Shwartz et~al.
\newblock Online learning and online convex optimization.
\newblock {\em Foundations and Trends{\textregistered} in Machine Learning},
  2012.

\bibitem{song1989study}
S. Song and B. Choi.
\newblock A study on continuous follow-the-leader (ftl) gaits: an effective
  walking algorithm over rough terrain.
\newblock {\em Mathematical biosciences}, 1989.

\bibitem{srinivasanclimb}
T Srinivasan, T Chang, L. Alva, G Chochlakis, M Rostami, and J Thomason.
\newblock Climb: A continual learning benchmark for vision-and-language tasks.
\newblock In {\em Thirty-sixth Conference on Neural Information Processing
  Systems Datasets and Benchmarks Track}.

\bibitem{tan2019mnasnet}
M. Tan, B. Chen, R. Pang, V. Vasudevan, M. Sandler, A. Howard, and Q. Le.
\newblock Mnasnet: Platform-aware neural architecture search for mobile.
\newblock In {\em Computer Vision and Pattern Recognition}, 2019.

\bibitem{thomee2016yfcc100m}
B. Thomee, D. Shamma, G. Friedland, B. Elizalde, K. Ni, D. Poland, D. Borth,
  and L. Li.
\newblock Yfcc100m: The new data in multimedia research.
\newblock {\em Communications of the ACM}, 2016.

\bibitem{vaswani2017attention}
A. Vaswani, N. Shazeer, N. Parmar, J. Uszkoreit, L. Jones, A. Gomez, {\L}.
  Kaiser, and I. Polosukhin.
\newblock {Attention Is All You Need}.
\newblock In {\em Neural Information Processing Systems}, 2017.

\bibitem{wang2023boosting}
Haonan Wang, Minbin Huang, Runhui Huang, Lanqing Hong, Hang Xu, Tianyang Hu,
  Xiaodan Liang, and Zhenguo Li.
\newblock Boosting visual-language models by exploiting hard samples.
\newblock {\em arXiv preprint arXiv:2305.05208}, 2023.

\bibitem{xie2018snas}
S. Xie, H. Zheng, C. Liu, and L. Lin.
\newblock {SNAS}: {Stochastic Neural Architecture Search}.
\newblock In {\em International Conference on Learning Representations}, 2019.

\bibitem{yan2022generative}
Shipeng Yan, Lanqing Hong, Hang Xu, Jianhua Han, Tinne Tuytelaars, Zhenguo Li,
  and Xuming He.
\newblock Generative negative text replay for continual vision-language
  pretraining.
\newblock In {\em European Conference on Computer Vision}, pages 22--38.
  Springer, 2022.

\bibitem{yao2022filip}
L. Yao, R. Huang, L. Hou, G. Lu, M. Niu, H. Xu, X. Liang, Z. Li, X. Jiang, and
  C. Xu.
\newblock Filip: Fine-grained interactive language-image pre-training.
\newblock In {\em International Conference on Learning Representations}, 2022.

\bibitem{you2022learning}
Haoxuan You, Luowei Zhou, Bin Xiao, Noel Codella, Yu Cheng, Ruochen Xu, Shih-Fu
  Chang, and Lu Yuan.
\newblock Learning visual representation from modality-shared contrastive
  language-image pre-training.
\newblock In {\em European Conference on Computer Vision}, pages 69--87.
  Springer, 2022.

\bibitem{2019Large}
Y. You, J. Li, S. Reddi, J. Hseu, S. Kumar, S. Bhojanapalli, X. Song, J.
  Demmel, K. Keutzer, and C. Hsieh.
\newblock Large batch optimization for deep learning: Training bert in 76
  minutes.
\newblock In {\em International Conference on Learning Representations}, 2020.

\bibitem{yu2021ernie}
F. Yu, J. Tang, W. Yin, Y. Sun, H. Tian, H. Wu, and H. Wang.
\newblock Ernie-vil: Knowledge enhanced vision-language representations through
  scene graphs.
\newblock In {\em Proceedings of the AAAI Conference on Artificial
  Intelligence}, volume~35, pages 3208--3216, 2021.

\bibitem{yu2022coca}
J. Yu, Z. Wang, V. Vasudevan, L. Yeung, M. Seyedhosseini, and Y. Wu.
\newblock Coca: Contrastive captioners are image-text foundation models.
\newblock Preprint arXiv:2205.01917, 2022.

\bibitem{zhai2022scaling}
X. Zhai, A. Kolesnikov, N. Houlsby, and L. Beyer.
\newblock Scaling vision transformers.
\newblock In {\em Proceedings of the IEEE/CVF Conference on Computer Vision and
  Pattern Recognition}, pages 12104--12113, 2022.

\bibitem{zoph2018learning}
B. Zoph, V. Vasudevan, J. Shlens, and Q. Le.
\newblock Learning {Transferable Architectures for Scalable Image Recognition}.
\newblock In {\em Computer Vision and Pattern Recognition}, 2018.

\end{thebibliography}
}

\end{document}